\pdfoutput=1

\documentclass[11pt]{article}

\usepackage[]{acl}

\usepackage{colortbl}
\usepackage{times}
\usepackage{latexsym}
\usepackage{booktabs}
\usepackage{multirow}
\usepackage{graphicx}
\usepackage{makecell}
\usepackage{bm}
\usepackage{amssymb}
\usepackage{amsmath}
\usepackage{arydshln}
\usepackage{pifont}

\usepackage[T1]{fontenc}

\usepackage[utf8]{inputenc}

\usepackage{microtype}

\usepackage{inconsolata}

\title{Distilling Fine-grained Sentiment Understanding from Large Language Models}

\author{Yice Zhang$^{1,3}$\thanks{\quad The first two authors contribute equally to this work.}\ , Guangyu Xie$^{1,3\ast}$, Hongling Xu$^{1,3}$, Kaiheng Hou$^{1}$,\\ 
\bf Jianzhu Bao$^{1,3}$, Qianlong Wang$^{1,3}$, Shiwei Chen$^{1,2}$, and Ruifeng Xu$^{1,2,3}$\thanks{\quad Corresponding Authors}\\
 $^{1}$ Harbin Institute of Technology, Shenzhen, China \\
 $^{2}$ Peng Cheng Laboratory, Shenzhen, China \\
 $^{3}$ Guangdong Provincial Key Laboratory of Novel Security Intelligence Technologies \\
\texttt{zhangyc\_hit@163.com,guangyuxie2001@gmail.com,xuruifeng@hit.edu.cn} \\
}

\begin{document}
\maketitle
\begin{abstract}

Fine-grained sentiment analysis (FSA) aims to extract and summarize user opinions from vast opinionated text.
Recent studies demonstrate that large language models (LLMs) possess exceptional sentiment understanding capabilities.
However, directly deploying LLMs for FSA applications incurs high inference costs.
Therefore, this paper investigates the distillation of fine-grained sentiment understanding from LLMs into small language models (SLMs).
We prompt LLMs to examine and interpret the sentiments of given reviews and then utilize the generated content to pretrain SLMs.
Additionally, we develop a comprehensive FSA benchmark to evaluate both SLMs and LLMs. Extensive experiments on this benchmark reveal that:
(1) distillation significantly enhances the performance of SLMs in FSA tasks, achieving a 6.00\% improvement in $F_1$-score, and the distilled model can outperform Llama-2-7b with only 220M parameters;
(2) distillation equips SLMs with excellent zero-shot sentiment classification capabilities, 
enabling them to match or even exceed their teacher models.
These results suggest that distillation from LLMs is a highly promising direction for FSA.\footnote{We will release our code, data, and pretrained model weights at \url{https://github.com/HITSZ-HLT/FSA-Distillation}.}
\end{abstract}

\section{Introduction}

\begin{figure}
\centering
\includegraphics[width=1.\linewidth]{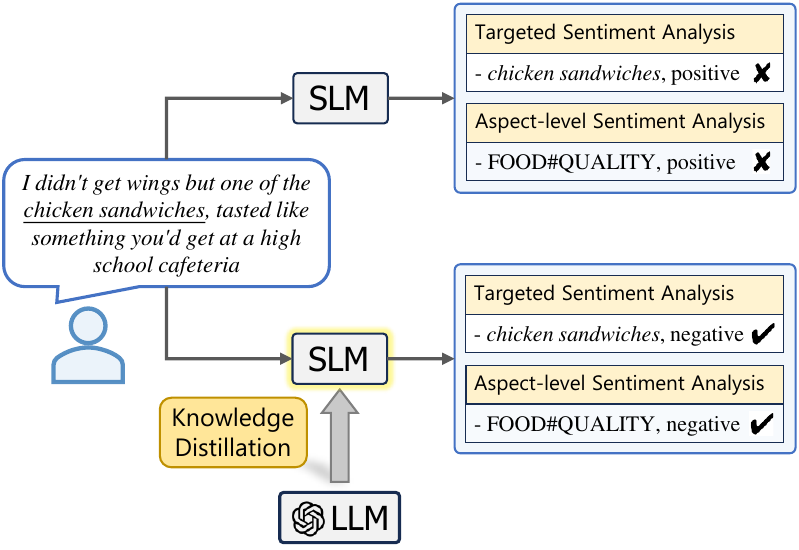}
\caption{Knowledge distillation from LLMs enhances SLMs' capabilities in handling complex sentiment contexts.}
\label{fig:intro}
\end{figure}

Fine-grained sentiment analysis (FSA) aims to thoroughly mine and understand user opinions from vast opinionated texts.
The two typical tasks of FSA are targeted sentiment analysis and aspect-level sentiment analysis, which organize and structure user opinions from the perspective of opinion targets\footnote{
Following \citet{pontiki-etal-2015-semeval,pontiki-etal-2016-semeval,li-etal-2018-transformation,Wan_Yang_Du_Liu_Qi_Pan_2020}, we adopt the term `opinion target' instead of `aspect term' to denote the reviewed entity in the text.
} and aspect categories, respectively.
These tasks are illustrated in Figure \ref{fig:intro}.

Existing FSA methods generally finetune small language models (SLMs) on labeled data \cite{xu-etal-2019-bert,Mao_Shen_Yu_Cai_2021,yan-etal-2021-unified,zhang-etal-2021-towards-generative}. 
Despite commendable results, these methods still face challenges when dealing with 
complex scenarios \cite{jiang-etal-2019-challenge,li-etal-2021-learning-implicit,fei-etal-2023-reasoning}. 
For example, the review depicted in Figure \ref{fig:intro} indirectly expresses negative feelings about the food through metaphorical language. However, methods relying on SLMs fail to recognize this negative sentiment as the context lacks explicit opinion words.

Recent studies have shown that large language models (LLMs) possess exceptional capabilities in natural language understanding \cite{qin-etal-2023-chatgpt,jiang2024mixtral}. Our observations also indicate that LLMs excel in understanding sentiment, as they can perform comprehensive and accurate sentiment analysis along with convincing interpretations.
For example, in the case depicted in Figure \ref{fig:intro}, LLMs interpret the review as suggesting: 
``\textit{Comparing the taste of the chicken sandwich to high school cafeteria food typically implies poor quality and lack of flavor, which is generally viewed negatively.}''
However, directly deploying LLMs for FSA applications incurs high computational costs due to their extensive parameters.
Additionally, it is a serious issue that LLMs often generate outputs that do not align with task-specific requirements.

This paper investigates the distillation of fine-grained sentiment understanding from LLMs into SLMs. The main advantage of this distillation process is that it allows us to leverage the exceptional capabilities of LLMs without incurring their high inference costs. Our research is structured around two core parts. 
\textbf{Firstly}, we present a formal definition of fine-grained sentiment understanding and, based on this definition, devise two types of prompts: an \textit{analysis} prompt and a \textit{rewriting} prompt. These prompts guide LLMs in generating content that embodies advanced sentiment understanding. Subsequently, we pretrain SLMs using this generated content to enhance their capabilities in sentiment analysis.
\textbf{Secondly}, we develop a comprehensive FSA benchmark for LLMs and SLMs.
For LLMs, we introduce a human evaluation framework that includes six dimensions, assessing both analysis results and reasoning processes. 
For SLMs, we refine existing FSA datasets to improve the effectiveness and reliability of the evaluations. 
Specifically, we identify `hard' samples in the existing datasets and further augment the datasets by annotating additional hard samples.

We conduct extensive experiments on our FSA benchmark and primarily arrive at the following conclusions.
(1) LLMs perform poorly in the FSA datasets with in-context learning but show exceptional improvement after finetuning, with an advantage of up to 8.31\% in $F_1$-score over SLMs.
(2) Distillation significantly enhances the performance of SLMs, achieving a 6.00\% $F_1$-score improvement, and the distilled model outperforms Llama-2-7b with only 220M parameters.
(3) Distillation enables SLMs to reach or even surpass their teacher models in zero-shot sentiment classification.
(4) The quality of the teacher model and the quantity of distillation data are two important factors affecting the effectiveness of distillation, with the latter being more influential.

\begin{figure}
\centering
\includegraphics[width=.99\linewidth]{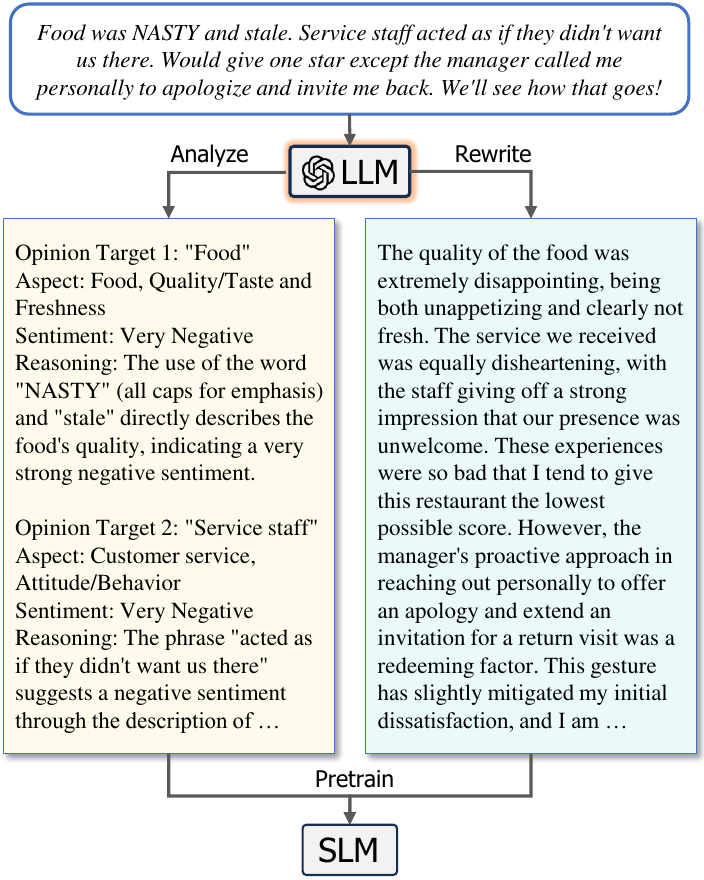}
\caption{Illustration of our distillation process.}
\label{fig:methods}
\end{figure}

\section{Distillation}
\label{sec:distillation}

\subsection{Prompting}
\label{sec:prompting}

Building upon \citet{pontiki-etal-2015-semeval,pontiki-etal-2016-semeval}, we define fine-grained sentiment understanding as a quadruple consisting of opinion target, aspect, sentiment, and reasoning. Specifically, (1) the opinion target refers to the reviewed entity within the text; (2) the aspect refers to the specific facets and dimensions of the opinion target being evaluated; (3) the sentiment indicates the strength of the sentiment toward the opinion target, ranging from very negative to very positive; and (4) reasoning refers to the process of inferring sentiment based on the context related to the opinion target.
The component of reasoning is central to this definition, as we believe that explicitly inferring sentiment from sentiment expressions or other indirect languages is essential for demonstrating a model's capability to understand sentiment.

Based on this definition, we develop two types of prompts:
(1) the analysis prompt is designed to guide LLMs in analyzing the review, listing the opinion targets, along with their corresponding aspect, sentiment, and reasoning;
(2) the rewriting prompt is intended to instruct LLMs to rewrite the review from a first-person perspective, expressing the user's feelings more clearly and evidently.
In the rewriting prompt, we require LLMs to incorporate the evaluated aspect, highlight the sentiment, and explicitly state the reasons for the sentiment in the rewritten review.
The main difference between these two prompts is that the analysis prompt produces structured outputs, while the rewriting prompt outputs natural language text.
As shown in Figure \ref{fig:methods}, these two prompts enable LLMs to examine and interpret the sentiments within the given review, generating clear and meaningful text from two distinct perspectives.
Complete prompt templates are presented in Appendix \ref{app:analysis-rewriting-prompt}.

\begin{table}[t]
\centering
\fontsize{8.5pt}{0.8\baselineskip}\selectfont
\begin{tabular}{l cccc} 
\toprule
\multirow{2}*{\textbf{Datasets}} & \multicolumn{2}{c}{\textbf{Yelp}} & \multicolumn{2}{c}{\textbf{Amazon}} \\
\cmidrule(lr){2-3} \cmidrule(lr){4-5}
& \textsc{Anl} & \textsc{Rw} & \textsc{Anl} & \textsc{Rw}\\
\midrule
{Llama-2-7b}   & 500K & 500K & 500K & 500K \\
{Mixtral-8x7b} & 500K & 500K & 500K & 500K \\
{GPT-3.5}      & 50K  & 50K  & 50K  & 50K  \\
\bottomrule
\end{tabular}
\caption{
Statistics of sentiment understanding corpus. 
\textsc{Anl} and \textsc{Rw} denote the analysis and rewriting prompts.
Due to budgetary constraints, we limit the usage of GPT-3.5 to 100K reviews, yielding 200K sentiment understanding texts.
}
\label{tab:sentiment-understanding-corpus-stat}
\end{table}

Using these two prompts, we construct a large-scale sentiment understanding corpus. Firstly, we collect one million reviews from the Yelp dataset\footnote{\url{https://www.yelp.com/dataset}} and the Amazon laptop dataset\footnote{\url{https://nijianmo.github.io/amazon/index.html}}, each accompanied by user ratings. We observe that reviews with mid-range ratings tend to exhibit more diverse sentiments, which 
motivates us to increase the proportion of such reviews during collection.
Secondly, we select three typical LLMs, including a proprietary GPT-3.5\footnote{Available at \url{https://chat.openai.com/}. The specific model used is \texttt{gpt-3.5-turbo-0125}.} and two open-source models, Llama-2-7b \cite{touvron2023llama} and Mixtral-8x7b \cite{jiang2024mixtral}.
Thirdly, we leverage the two proposed prompts to guide these models in generating sentiment understanding texts for these reviews, thereby constructing a sentiment understanding corpus. The statistical information of this corpus is presented in Table \ref{tab:sentiment-understanding-corpus-stat}.

\subsection{Pretraining}

We use T5 \cite{JMLR:v21:20-074} as the default SLM and continue to pretrain it using \textsc{seq2seq} objective on our sentiment understanding corpus. This corpus comprises reviews and corresponding sentiment understanding texts. During pretraining, these reviews serve as inputs, and the sentiment understanding texts serve as targets.
The pretraining loss can be formulated as follows:
\begin{equation}
    L = -\sum_t \log P(\bm u_t|\bm x,\bm u_{<t}),
\end{equation}
where $\bm x$ and $\bm u$ denote the input review and the sentiment understanding text, respectively.
We claim that such pretraining can increase the sensitivity of SLMs toward sentiment elements and enhance their capabilities to understand the context within opinionated texts.

Our pretraining approach is similar to two existing methods. The first, domain-adaptive pretraining \cite{xu-etal-2019-bert,gururangan-etal-2020-dont}, continues to train models on domain-specific texts using language modeling objectives. 
This method is self-supervised, whereas our pretraining utilizes explicit supervisory signals from LLMs.
The second, sentiment-enhanced pretraining \cite{tian-etal-2020-skep,zhang-etal-2023-empirical}, enhances the pretraining process using sentiment knowledge such as sentiment lexicons and user ratings. A typical technique is rating prediction \cite{zhou-etal-2020-sentix,li-etal-2021-learning-implicit}, which encourages models to learn sentiment-aware representations by predicting the rating scores of reviews. Our pretraining, in contrast, incorporates more fine-grained and comprehensive supervisory signals.

\section{FSA Benchmark}
\label{sec:fsa-benckmark}

During the distillation process, we encounter several critical questions: (a) how to measure the sentiment understanding capabilities of LLMs when selecting a teacher; (b) what is the gap between the student and the teacher?; and (c) how much is this gap reduced after distillation? 
The prerequisite for answering these questions is that we can conduct effective and reliable evaluations of different models.
To this end, we develop a fine-grained sentiment analysis (FSA) benchmark.

\vspace{6pt}
\noindent
\textbf{Tasks.}
This benchmark includes two FSA tasks: targeted sentiment analysis and aspect-level sentiment analysis. Targeted sentiment analysis aims to extract opinion targets from the text and determine their sentiment polarities. Aspect-level sentiment analysis aims to detect aspect categories evaluated in the text and determine their sentiment polarities.

In these two tasks, we particularly concentrate on model performance in complex contexts. 
Our focus is directed toward implicit and multiple sentiments, where prior studies suggest that SLMs exhibit subpar performance \citep{jiang-etal-2019-challenge, li-etal-2021-learning-implicit}.
Implicit sentiments are those not directly expressed through explicit opinion words but inferred indirectly through context, including factual statements, comparisons, metaphors, etc.
Multiple sentiments occur when a sentence expresses different sentiment polarities toward different opinion targets or aspect categories.

\vspace{6pt}
\noindent
\textbf{Datasets.}
We derive base FSA datasets from SemEval Challenges \cite{pontiki-etal-2014-semeval, pontiki-etal-2016-semeval} and make two key additions. 
Firstly, we identify `hard' samples in the test set of these datasets, namely those containing implicit or multiple sentiments.
Here, we use opinion words to determine whether a sample contains implicit sentiment. The opinion word annotations for SemEval-14 datasets are sourced from \citet{fan-etal-2019-target}. For SemEval-16 datasets, we annotate the opinion words for each opinion target and aspect category.

Secondly, we annotate additional hard samples. 
We observe that the original test sets contain too few hard samples, which reduces the reliability of the evaluations.
Therefore, we view 300 reviews for each dataset and annotate those sentences containing implicit or multiple sentiments. 
The annotation process is described in Appendix \ref{app:fsa-annotation}. 
We then supplement these samples into the original test sets. The statistics of these datasets are listed in Table \ref{tab:fsa-stat}.

\begin{table}[t]
\centering
\fontsize{8.5pt}{0.8\baselineskip}\selectfont
\setlength\tabcolsep{3.73pt}
\begin{tabular}{l cc c cc c cc} 
\toprule
\textbf{Datasets} & \textbf{Split} & \textbf{\#Sent} & \textbf{\#Trg} & \textbf{\#Asp} & \textbf{\#Imp} & \textbf{\#Mul} \\
\midrule
\multirow{3}*{{TSA-Rest14}} & Train & 2432 & 2972 & - & -   & 277 \\
                                   & Dev   & 609  & 721  & - & -   & 78 \\
                                   & Test  & 800  & 1134 & - & 192 & 85 \\
\midrule
\multirow{3}*{{TSA-Laptop14}} & Train & 2436 & 1922 & - & -   & 148 \\
                                     & Dev   & 609  & 436  & - & -   & 34 \\
                                     & Test  & 800  & 654  & - & 133 & 40 \\
\midrule
\multirow{3}*{{ASA-Rest16}} & Train & 1600 & 1386 & 1823 & -   & 114 \\
                                   & Dev   & 400  & 386  & 477  & -   & 29 \\
                                   & Test  & 676  & 623  & 751  & 199 & 42 \\
\midrule
\multirow{3}*{{ASA-Laptop16}} & Train & 2000 & - & 2349 & -   & 126 \\
                                     & Dev   & 500  & - & 560  & -   & 25 \\
                                     & Test  & 808  & - & 801  & 250 & 35 \\
\midrule
\multirow{1}*{{Rest-Hard}}   & Test  & 340  & 383  & 504  & 285 & 104 \\
\midrule
\multirow{1}*{{Laptop-Hard}} & Test  & 237  & 290 & 382  & 212 & 59 \\
\bottomrule
\end{tabular}
\caption{
Statistics of FSA datasets, including two targeted sentiment analysis (TSA) datasets and two aspect-level sentiment analysis (ASA) datasets. 
\#Sent, \#Trg, and \#Asp denote the number of sentences, targets, and aspects.
\#Imp and \#Mul denote the number of samples with implicit and multiple sentiments.
}
\label{tab:fsa-stat}
\end{table}

\vspace{6pt}
\noindent
\textbf{Evaluation Framework for LLMs.}
Using the above FSA datasets to evaluate SLMs is effective, but they are inadequate and restrictive for evaluating LLMs. There are two main reasons.
Firstly, these datasets do not assess the reasoning process, which is crucial to thoroughly evaluating the sentiment understanding capabilities of LLMs. Secondly, LLMs frequently generate outputs that deviate from the task-specific requirements, preventing them from fully demonstrating their capabilities. While finetuning can tackle this issue for open-source models, it is often impractical for proprietary models.

Therefore, we introduce an evaluation framework for LLMs. Initially, we leverage the analysis prompt outlined in Section \ref{sec:prompting} to guide LLMs in articulating their sentiment understanding in quadruple form. We then systematically evaluate each component of these quadruples: 
(1) \textbf{precision} and \textbf{recall} for target-aspect assess the correctness of identified target-aspects and check whether any mentioned in the review are omitted;
(2) \textbf{accuracy} of sentiment quantifies the correctness of the assigned sentiment strengths;
(3) \textbf{persuasiveness} and \textbf{exhaustiveness} of reasoning examine whether the reasoning processes convincingly support the sentiments and comprehensively enumerate relevant context;
(4) \textbf{hallucination} in reasoning investigates whether LLMs refer to non-existent segments of the review or misattribute descriptions to wrong subjects.
We formulate detailed annotation guidelines and have humans evaluate each dimension with scores of 0-2 points.
The annotation guidelines and the evaluation process are presented in Appendix \ref{app:human-evaluation}.

\section{Experiments}
\subsection{Experimental Setup}

\textbf{Implementation Details.}
In the distillation process, we select T5-base as the student model and Llama-2-7b, Mixtral-8x7b, and GPT-3.5 as the teacher models. 
For each teacher model, we construct four datasets by applying the analysis and rewriting prompts on Yelp and Amazon data\footnote{For GPT-3.5, we utilize OpenAI's API to generate sentiment understanding texts. For Llama-2-7b, these texts are generated on a single 48G A6000 GPU using the vLLM \cite{10.1145/3600006.3613165} framework. For Mixtral-8x7b, text generation is conducted on four A6000 GPUs, also utilizing the vLLM framework.
}, the statistics of which are presented in Table \ref{tab:sentiment-understanding-corpus-stat}. 
During pretraining, we merge these four datasets into a single corpus to pretrain the T5 model.
By default, our distillation process utilizes 100K reviews. We set the batch size to 100, the number of training epochs to 10, and the initial learning rate to 3e-3.
Additional hyperparameters are detailed in Appendix \ref{app:implementation-details}.

After pretraining, we evaluate the distilled models on four FSA datasets, the statistics of which are provided in Table \ref{tab:fsa-stat}. During the evaluation, we merge the additionally annotated `hard' samples into the original test set: Rest-Hard for TSA-Rest14 and ASA-Rest16, and Laptop-Hard for TSA-Laptop14 and ASA-Laptop16. 
To minimize the impact of randomness, we run all finetuning experiments 10 times and report the average results.

\vspace{6pt}
\noindent
\textbf{Baselines.} We compare our method against three types of baselines. The first is a strong FSA method: \textsc{InstructABSA} \cite{scaria-etal-2024-instructabsa}, which conducts instruction tuning using task definition and positive examples. We reproduce this method on our FSA datasets using the backbone of \texttt{tk-instruct-base-def-pos} \cite{wang-etal-2022-super}. The second baseline consists of classic pretraining methods for sentiment analysis, including domain adaptation pretraining (DAPT) \cite{xu-etal-2019-bert,gururangan-etal-2020-dont} and rating prediction \cite{zhou-etal-2020-sentix,li-etal-2021-learning-implicit}. Finally, we compare our approach with a general distillation method. For this, we pretrain T5-base using a large instruction dataset \cite{wu-etal-2024-lamini}, which contains 2.58 million diverse instruction samples generated using GPT-3.5. In addition, we also provide results of three LLMs using in-context learning and supervised fine-tuning for reference.

\subsection{Evaluation of LLMs}

We first evaluate the sentiment understanding capabilities of several LLMs and compare their performance against SLMs.

\subsubsection{Human Evaluation}
\begin{figure}[t]
\centering
\includegraphics[width=0.88\linewidth]{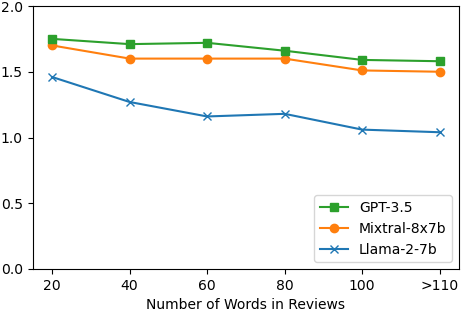}
\caption{
The trend of LLMs' human evaluation scores with varying review lengths  (scoring range 0-2 points).
}
\label{fig:llm-evaluation-length}
\end{figure}

We conduct human evaluations of the three selected LLMs using the evaluation framework outlined in the \textit{FSA Benchmark} section. 
The results are presented in Table \ref{tab:llm-human-evaluation}.
We make the following observations. Firstly, Llama-2-7b performs notably worse than the other two LLMs, especially with missing target-aspects, less persuasive reasoning, and more hallucinations.
Secondly, the performance of Mixtral-8x7b and GPT-3.5 is comparable and generally good. Thirdly, in the \textsc{laptop} domain, LLMs are more prone to produce incorrect target-aspects, which can be attributed to the greater diversity of target-aspects in this domain. For example, in the FSA datasets, the number of aspect categories for \textsc{restaurant} is 12, whereas for \textsc{laptop}, it exceeds 80.
In addition, we examine the performance of LLMs on reviews of different lengths. As shown in Figure \ref{fig:llm-evaluation-length}, the longer the review, the poorer the performance. This trend is particularly pronounced with Llama-2-7b, indicating its weak capability in handling long reviews.

\begin{table}[t]
\centering
\fontsize{8.35pt}{0.8\baselineskip}\selectfont
\setlength\tabcolsep{0.2pt}
\begin{tabular}{p{1.6cm} >{\centering\arraybackslash}p{.85cm} >{\centering\arraybackslash}p{.85cm} >{\centering\arraybackslash}p{.85cm} >{\centering\arraybackslash}p{.85cm} >{\centering\arraybackslash}p{.85cm} > 
{\centering\arraybackslash}p{.85cm} >{\raggedleft\arraybackslash}p{.7cm}}
\toprule
\multirow{2}*{\textbf{Methods}} & \multicolumn{2}{c}{\textbf{Target-Aspect}} & \multicolumn{1}{c}{\textbf{Senti}} & \multicolumn{3}{c}{\textbf{Reasoning}} & \multirow{2}*{\textbf{Avg}} \\
\cmidrule(lr){2-3} \cmidrule(lr){4-4} \cmidrule(lr){5-7}
& \texttt{Prec} & \texttt{Rec} & \texttt{Accu} 
& \texttt{Pers} & \texttt{Exha} & \texttt{Hall} &  \\
\midrule
& \multicolumn{6}{c}{\textsc{Restaurant Domain}} \\
\hdashline[2pt/4pt]
Llama-2-7b & 1.35 & 0.80 & 1.09 & 1.34 & 0.68 & 1.27 & 1.09\\ 
Mixtral-8x7b & 1.79 & 1.40 & 1.56 & 1.62 & 1.18 & 1.87 & 1.57\\ 
GPT-3.5 & 1.88 & 1.56 & 1.54 & 1.69 & 1.45 & 2.0 & 1.69 \\ 
\midrule
& \multicolumn{6}{c}{\textsc{Laptop Domain}} \\
\hdashline[2pt/4pt]
Llama-2-7b & 1.21 & 1.48 & 0.69 & 1.41 & 1.70 & 1.91 & 1.40 \\ 
Mixtral-8x7b & 1.40 & 1.74 & 1.17 & 1.85 & 1.78 & 1.95 & 1.65 \\ 
GPT-3.5 & 1.43 & 1.85 & 1.15 & 1.87 & 1.83 & 1.97 & 1.69 \\ 
\bottomrule
\end{tabular}
\caption{
Human scores of LLMs for fine-grained sentiment understanding (scoring range 0-2 points).
`Senti' refers to sentiments.
\texttt{Pers}, \texttt{Exha}, and \texttt{Hall} denote persuasiveness, exhaustiveness, and hallucination.
}
\label{tab:llm-human-evaluation}
\end{table}

\begin{table*}[t]
\centering
\fontsize{8.5pt}{0.8\baselineskip}\selectfont
\setlength\tabcolsep{2.88pt}
\begin{tabular}{l c ccc c ccc c ccc c ccc c c} 
\toprule
\multirow{2}*{\textbf{Methods}} && \multicolumn{3}{c}{\textbf{TSA-Rest14}} && \multicolumn{3}{c}{\textbf{TSA-Laptop14}} && \multicolumn{3}{c}{\textbf{ASA-Rest16}} && \multicolumn{3}{c}{\textbf{ASA-Laptop16}} && \multirow{2}*{\textbf{Avg}} \\
\cmidrule(lr){3-5}  \cmidrule(lr){7-9} \cmidrule(lr){11-13} \cmidrule(lr){15-17}
&& \texttt{All} & \texttt{Imp} & \texttt{Mul} 
&& \texttt{All} & \texttt{Imp} & \texttt{Mul}  
&& \texttt{All} & \texttt{Imp} & \texttt{Mul}  
&& \texttt{All} & \texttt{Imp} & \texttt{Mul} \\
\midrule
T5-base && 67.33 & 51.73 & 57.23 && 60.78 & 50.40 & 53.08 && 68.70 & 62.11 & 63.66 && 49.36 & 43.32 & 46.70 && 56.20\\
\textsc{InstructABSA} && 67.88 & 52.11 & 55.89 && 62.94 & 51.61 & 54.09 && 71.49 & 66.17 & 63.10 && 49.77 & 42.76 & 49.39 && +1.05\\
DAPT &&  68.31 & 52.77 & 58.80 && 61.42 & 51.00 & 53.77 && 71.31 & 66.01 & 65.28 && 50.23 & 44.06 & 46.03 &&+1.22\\
\textsc{Rating Prediction} && 69.35 & 54.65 & 59.77 && 61.09 & 50.06 & 51.31 && 70.53 & 64.41 & 65.04 && 50.56 & 44.19 & 45.29 && +0.99\\
\textsc{General Distillation} && 66.91 & 51.65 & 57.45 && 59.54 & 48.20 & 52.29 && 67.15 & 60.34 & 61.22	&& 46.91	& 39.86 & 45.35 && -1.46\\
\midrule
\multicolumn{4}{l}{\textbf{Llama-2-7b}} \\
{\textsc{In-context Learning}} && 40.02 & 34.81 & 34.11 && 19.35 & 19.52 & 21.69 && 41.68 & 38.43 & 28.05 && 18.81 & 19.89 & 10.13 && -28.99\\ 
\textsc{Supervised Finetuning} && 70.72 & 58.47 & 61.81 && 63.39 & 54.16 & 55.94 && 75.44 & 70.03 & 71.03 && 57.55 & 51.60 & 55.25 && +5.92 \\ 
\hdashline[2pt/4pt]
\textsc{Distil (Anl w/o R)} && 68.17 & 52.94 & 60.14 && 62.08 & 51.56 & 54.33 && 70.24 & 64.82 & 64.79 && 50.47 & 44.65 & 48.73 && +1.54\\ 
\textsc{Distil (Anl w/o L)} && 68.74 & 53.98 & 60.26 && 62.71 & 52.61 & 54.34 && 70.72 & 64.55 & 63.62 && 51.21 & 44.87 & 47.51 && +1.73 \\ 
\textsc{Distil (Anl)} && 68.89 & 53.86 & 60.33 && 62.72 & 52.50 & 54.61 && 70.85 & 64.71 & 65.24 && 51.16 & 45.37 & 50.31 && +2.18\\ 
\textsc{Distil (Rw)} && 69.20 & 53.69 & 59.93 && 62.74 & 53.28 & 55.19 && 71.86 & 66.15 & 66.97 && 52.14 & 46.28 & 50.40 && +2.79\\ 
\rowcolor{gray!20} 
\textsc{Distil (Anl\&Rw)} && 68.85 & 54.44 & 60.64 && 63.21 & 53.16 & 55.10 && 72.27 & 66.38 & 66.13 && 51.81 & 45.75 & 50.30 && +2.80 \\ 
\rowcolor{gray!20} 
\textsc{Distil (Anl\&Rw, 1M)} && 69.96 & 56.17 & 61.15 && 63.93 & 54.25 & 57.16 && 74.17 & 68.47 & 70.00 && 54.70 & 49.14 & 53.43 && +4.84\\ 
\midrule
\multicolumn{4}{l}{\textbf{Mixtral-8x7b}} \\
{\textsc{In-context Learning}} && 51.15 & 43.15 & 45.42 && 30.74 & 32.00 & 39.29 && 54.87 & 49.83 & 50.91 && 28.68 & 26.52 & 29.48 && -16.03 \\ 
\textsc{Supervised Finetuning} && 72.59 & 59.97 & 64.21 && 64.40 & 55.34 & 55.95 && 78.35 & 74.03 & 73.46 && 60.25 & 55.45 & 60.18 && +8.31 \\ 
\hdashline[2pt/4pt]
\textsc{Distil (Anl w/o R)} && 68.12 & 53.04 & 59.49 && 62.92 & 52.66 & 56.82 && 69.21 & 63.05 & 64.18 && 50.90 & 45.14 & 50.54 && +1.81 \\ 
\textsc{Distil (Anl w/o L)} && 69.29 & 54.63 & 61.76 && 62.97 & 53.02 & 55.81 && 71.21 & 65.45 & 64.90 && 51.25 & 45.03 & 47.92 && +2.48\\ 
\textsc{Distil (Anl)} && 69.21 & 54.34 & 61.16 && 63.07 & 53.60 & 54.93 && 71.51 & 65.66 & 66.32 && 51.68 & 46.19 & 50.36 && +2.80 \\ 
\textsc{Distil (Rw)} && 68.95 & 54.00 & 60.22 && 62.21 & 51.27 & 53.40 && 71.96 & 66.02 & 65.92 && 52.16 & 46.42 & 50.45 && +2.38 \\ 
\rowcolor{gray!20} 
\textsc{Distil (Anl\&Rw)} && 69.69 & 54.96 & 61.82 && 63.77 & 53.63 & 57.43 && 72.13 & 66.60 & 66.92 && 52.94 & 47.30 & 51.91 && +3.73 \\ 
\rowcolor{gray!20} 
\textsc{Distil (Anl\&Rw, 1M)} && 71.21 & 58.08 & 62.48 && 65.04 & 55.01 & 59.07 && 75.33 & 70.07 & 71.34 && 55.30 & 49.05 & 54.42 && +6.00 \\ 
\midrule
\multicolumn{4}{l}{\textbf{GPT-3.5}} \\
\multicolumn{1}{l}{\textsc{In-context Learning}} && 51.28 & 43.11 & 48.35 && 28.69 & 30.73 & 35.07 && 56.63 & 52.75 & 52.23 && 28.90 & 29.25 & 32.54 && -15.41 \\ 
\hdashline[2pt/4pt]
\textsc{Distil (Anl w/o R)} && 68.86 & 53.81 & 59.70 && 62.45 & 51.65 & 54.85 && 71.29 & 65.76 & 65.97 && 51.28 & 45.30 & 47.68 && +2.02 \\ 
\textsc{Distil (Anl w/o L)} && 68.99 & 54.20 & 60.54 && 62.34 & 52.30 & 55.10 && 71.84 & 65.48 & 66.93 && 51.79 & 45.80 & 50.36 && +2.61\\ 
\textsc{Distil (Anl)} && 69.23 & 55.06 & 60.19 && 62.84 & 52.74 & 55.00 && 71.21 & 65.51 & 65.22 && 51.63 & 45.39 & 49.95 && +2.46 \\ 
\textsc{Distil (Rw)}&& 69.35 & 54.64 & 60.60 && 62.30 & 51.92 & 53.13 && 71.60 & 65.71 & 66.91 && 51.95 & 45.84 & 49.17 && +2.39 \\
\rowcolor{gray!20} 
\textsc{Distil (Anl\&Rw)}&& 69.45 & 55.28 & 61.03 && 62.96 & 52.96 & 56.66 && 72.80 & 67.06 & 68.11 && 53.11 & 47.49 & 52.19 && +3.73\\ 
\bottomrule
\end{tabular}
\caption{
Experimental results on four FSA datasets in fully-supervised settings ($F_1$-score, \%).
The results on the origin test sets are presented in Appendix \ref{app:additional-results}.
\texttt{Imp} and \texttt{Mul} denote implicit and multiple sentiments.
\textsc{Anl} and \textsc{Rw} denote the analysis and rewriting prompts.
\textsc{Anl w/o R} and \textsc{w/o L} indicate the removal of reasoning and labels (\textit{i.e.}, target, aspect, and sentiment) from the \textsc{Anl} corpus.
\textsc{1M} indicates distillation using 1 million reviews, while other results are based on 100K reviews.
}
\label{tab:fsa-supervised}
\end{table*}

\subsubsection{Performance on FSA Datasets}
We also finetune the two open-source LLMs on our FSA datasets for a more objective evaluation. As shown in Table \ref{tab:fsa-supervised}, LLMs significantly outperform SLMs, achieving average $F_1$-score improvements of 5.92\% for Llama-2-7b and 8.31\% for Mixtral-8x7b. 
These results indicate that LLMs have a substantial advantage over SLMs in FSA tasks.

\subsection{Evaluation of Distilled Models}

Next, we conduct experiments to evaluate how the distilled models perform in fully-supervised and zero-shot settings. 

\subsubsection{Fully-supervised Setting}
Table \ref{tab:fsa-supervised} lists the comparison results on the FSA datasets. 
According to these results, distillation significantly enhances the performance of SLMs, achieving up to a 6.00\% improvement in $F_1$-score. With Mixtral-8x7b as the teacher model, the distilled model outperforms Llama-2-7b with only 220M parameters.
In addition, our approach significantly outperforms baseline methods, \textit{i.e.}, \textsc{InstructABSA}, DAPT, \textsc{Rating Prediction}, and \textsc{General Distillation}. 
These findings demonstrate the effectiveness and substantial potential of our distillation approach.

Furthermore, we make the following observations.
(1) The distilled models achieve greater performance gains on implicit and multiple sentiments. 
(2) The quality of the teacher model has a significant impact: better teachers produce better students.
(3) The analysis and rewriting prompts are both effective for distillation, showing similar effects. 
Their combination yields the best results.
(4) The reasoning process in the analysis prompt is crucial, as removing the reasoning process significantly reduces the performance of the distilled models.
(5) Increasing the number of reviews for distillation from 100K to 1 million notably improves the performance of the distilled models, with an average $F_1$-score increase of over 2\%.

\begin{table*}[t]
\centering
\fontsize{8.5pt}{0.8\baselineskip}\selectfont
\setlength\tabcolsep{2.9pt}
\begin{tabular}{l c ccc c ccc c ccc c ccc cc} 
\toprule
\multirow{2}*{\textbf{Methods}}  && \multicolumn{3}{c}{\textbf{TSA-Rest14}} && \multicolumn{3}{c}{\textbf{TSA-Laptop14}} && \multicolumn{3}{c}{\textbf{ASA-Rest16}} && \multicolumn{3}{c}{\textbf{ASA-Laptop16}} && \multirow{2}*{\textbf{Avg}}\\
\cmidrule(lr){3-5}  \cmidrule(lr){7-9} \cmidrule(lr){11-13} \cmidrule(lr){15-17}
&& \texttt{All} & \texttt{Imp} & \texttt{Mul} 
&& \texttt{All} & \texttt{Imp} & \texttt{Mul}  
&& \texttt{All} & \texttt{Imp} & \texttt{Mul}  
&& \texttt{All} & \texttt{Imp} & \texttt{Mul} \\
\midrule
T5-base && 66.58 & 55.92 & 46.78 && 65.25 & 58.13 & 46.85 && 70.60 & 66.46 & 49.85 && 72.53 & 67.56 & 50.45 && 59.75 \\
\midrule
Llama-2-7b && 74.36 & 63.20 & 50.78 && 71.08 & 62.40 & 48.20 &&	77.77 & 75.80 & 48.00 && 80.56 & 77.62 & 47.75 && 64.79 \\
\rowcolor{gray!20} 
\textsc{Distil (Anl\&Rw, 1M)} && 78.11 & 67.76 & 58.54 && 76.48 & 70.73 & 62.16 && 67.41 & 59.11 & 44.62 && 73.54 & 69.69 & 51.35 && 64.96 \\
\midrule
Mixtral-8x7b && 83.85 & 76.18 & 70.51 && 79.13 & 73.78 & 64.41 && 85.58 & 82.70 & 64.31 && 85.71 & 82.01 & 65.32 && 76.12 \\
\rowcolor{gray!20} 
\textsc{Distil (Anl\&Rw, 1M)} && 83.45 & 74.75 & 70.07 && 81.04 & 75.61 & 70.72 && 84.14 & 80.86 & 67.38 && 85.55 & 81.73 & 71.62 && 77.24\\
\midrule
GPT-3.5 && 78.97 & 67.33 & 60.98 && 77.22 & 71.54 & 57.21 && 79.76 & 76.57 & 55.69 && 81.32 & 77.48 & 52.70 && 69.73 \\ 
\rowcolor{gray!20} 
\textsc{Distil (Anl\&Rw, 100K)} && 79.96 & 69.04 & 64.30 && 77.54 & 70.12 & 60.36 && 79.60 & 75.04 & 54.15 && 80.64 & 76.06 & 58.11 && 70.41 \\ 
\bottomrule
\end{tabular}
\caption{
Experimental results of zero-shot sentiment classification on four FSA datasets (accuracy, \%).
The prompts are presented in Appendix \ref{app:zero-shot-prompt}.
\texttt{Imp} and \texttt{Mul} denote implicit and multiple sentiments.
\textsc{Anl} and \textsc{Rw} denote the analysis and rewriting prompts.
}
\label{tab:fsa-zero-shot}
\end{table*}

\begin{table}[t]
\centering
\fontsize{8.5pt}{0.8\baselineskip}\selectfont
\setlength\tabcolsep{1.18pt}
\begin{tabular}{p{1.3cm} >{\centering\arraybackslash}p{1.25cm} >{\centering\arraybackslash}p{1.25cm} >{\centering\arraybackslash}p{1.25cm} >{\centering\arraybackslash}p{1.25cm} >{\raggedleft\arraybackslash}p{0.9cm}} 
\toprule
\textbf{Methods} & \textbf{Rest14} & \textbf{Laptop14} & \textbf{Rest16} & \textbf{Laptop16} & \textbf{Avg}\\

\midrule
T5-base            & 57.23 & 53.08 & 63.66 & 46.70 & 55.17 \\ 
\hdashline[2pt/4pt]
\textsc{R11111}    & 60.38 & 55.41 & 67.22 & {52.58} & +3.73\\
\textsc{R00100}    & 60.87 & 53.11 & 68.35 & 51.91 & +3.39\\
\textsc{R12421}    & {61.82} & {57.43} & 66.92 & 51.91 & +4.35\\ 
\bottomrule
\end{tabular}
\caption{
Effect of sampling schemes on samples with multiple sentiments ($F_1$-score, \%):
(1) R11111 represents sampling reviews according to their original proportions;
(2) R00100 indicates exclusively sampling 3-star reviews.
(3) R12421 denotes oversampling 2-star and 4-star reviews by a factor of 2, and 3-star reviews by a factor of 4.
The teacher model for this experiment is Mixtral-8x7b.
}
\label{tab:sampling-result}
\end{table}

During data collection, we increase the proportion of reviews with mid-range ratings, hypothesizing that these reviews generally exhibit more diverse sentiments and could enhance the model's capability to handle samples with multiple sentiments. We conduct experiments with different sampling schemes. The results in Table \ref{tab:sampling-result} show that the oversampling scheme (\textit{i.e.}, R12421) achieves the best overall performance, confirming our hypothesis.

\subsubsection{Zero-shot Setting}
Existing works indicate that LLMs excel in zero-shot sentiment classification, approaching the performance of finetuned SLMs \cite{zhang2023sentiment,wang2024chatgpt}. We expect that distillation could endow SLMs with similar capabilities. Surprisingly, the results in Table \ref{tab:fsa-zero-shot} reveal that distillation not only significantly improves the zero-shot sentiment classification performance of SLMs but also enables them to reach and even surpass their teacher models. 
This superior performance can be attributed to the fact that teacher models are generally designed for broader tasks, while student models are specifically trained on a large-scale sentiment-specific corpus, equipping them with stronger sentiment classification capabilities.
This finding suggests that distillation from LLMs is a highly promising approach in scenarios where training data is unavailable.

\subsection{Further Analysis}

\begin{figure}[t]
\centering
\includegraphics[width=1.\linewidth]{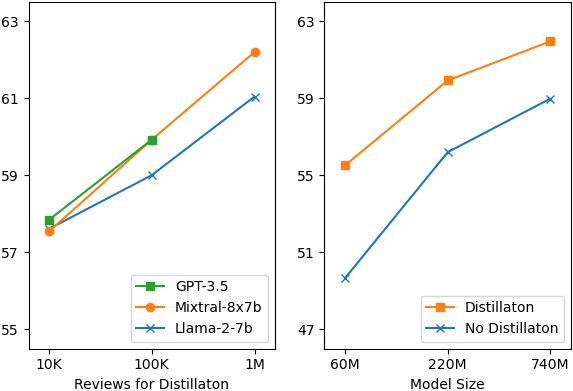}
\caption{Scaling trends of review quantity and model size (average $F_1$-score on FSA datasets, \%).
} 
\label{fig:scaling}
\end{figure}

\begin{figure}[t]
\centering
\includegraphics[width=0.77\linewidth]{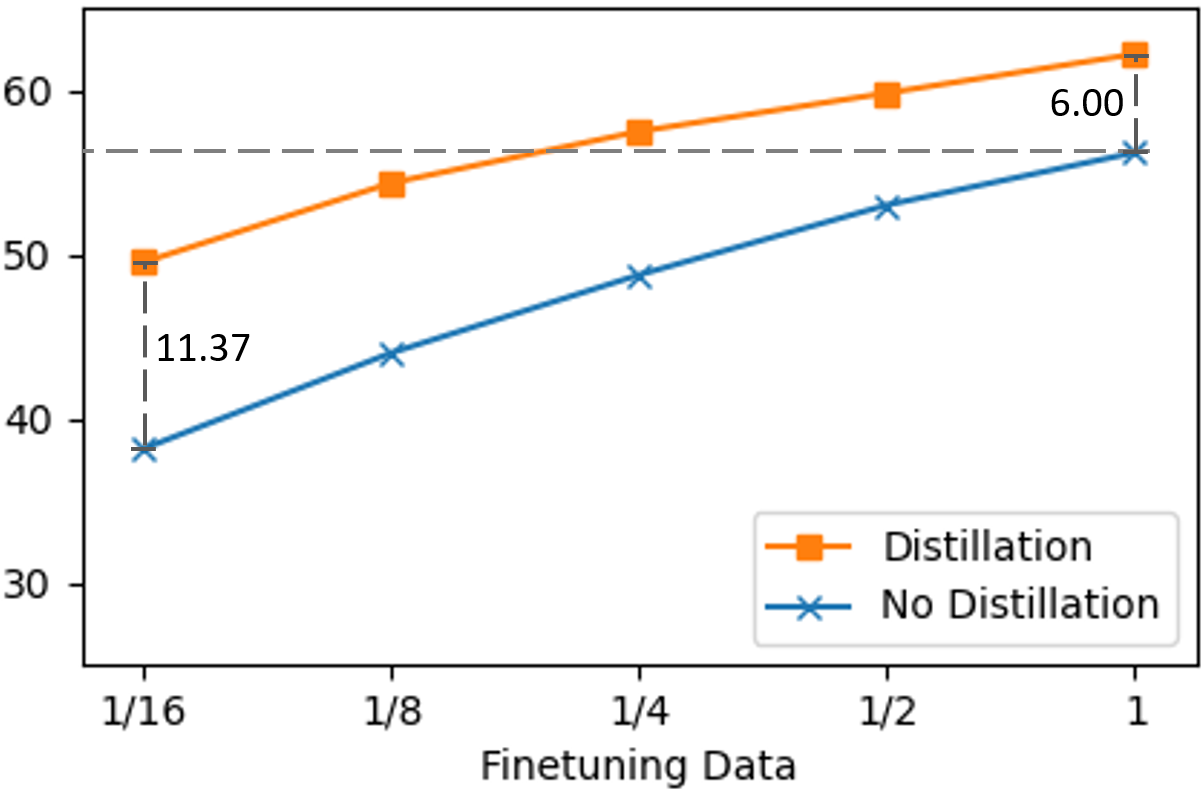}
\caption{
Performance on data-scarce scenarios (average $F_1$-score on FSA datasets, \%).
Here, the teacher model is Mixtral-8x7b, and the number of reviews for distillation is 1 million.
}
\label{fig:scarce}
\end{figure}

\subsubsection{Scaling Trends}
The results in Table \ref{tab:fsa-supervised} demonstrate that increasing the number of reviews notably improves the effectiveness of distillation.
We further investigate the impact of review quantity.
Figure \ref{fig:scaling} illustrates that the performance of the distilled models steadily improves as the number of reviews increases, highlighting the potential benefits of further data expansion. 
Moreover, we find that the quantity of reviews has a greater impact than the quality of the teacher model. 
Even though Llama-2-7b is markedly inferior to Mixtral-8x7b and GPT-3.5, when distilled with 1 million reviews, its student model can outperform those of the latter two distilled with 100K reviews.

\begin{table*}[t]
\centering
\fontsize{8.5pt}{0.8\baselineskip}\selectfont
\setlength\tabcolsep{3pt}
\begin{tabular}{l cccc cccc cccc ccc} 
\toprule
\multirow{2}*{\textbf{Methods}}  & \multicolumn{3}{c}{\textbf{TSA-Rest14}} && \multicolumn{3}{c}{\textbf{TSA-Laptop14}} && \multicolumn{3}{c}{\textbf{ASA-Rest16}} && \multicolumn{3}{c}{\textbf{ASA-Laptop16}} \\
\cmidrule(lr){2-4} \cmidrule(lr){6-8} \cmidrule(lr){10-12} \cmidrule(lr){14-16}
& \texttt{Type1} & \texttt{Type2} & \texttt{Type3} && \texttt{Type1} & \texttt{Type2} & \texttt{Type3} && \texttt{Type1} & \texttt{Type2} & \texttt{Type3} && \texttt{Type1} & \texttt{Type2} & \texttt{Type3}  \\
\midrule
T5-base & 14.94 & 3.08 & 18.03 && 11.40 & 1.04 & 21.76 && 15.46 & 8.37 & 19.96 && 13.18 & 7.32 & 34.40 \\
\hdashline[2pt/4pt]
Mixtral-8x7b (\textsc{Finetuning}) & 20.40 & 2.85 & 11.86 && 13.64 & 0.86 & 15.89 && 19.10 & 6.65 & 6.01 && 19.40 & 6.59 & 23.79 \\
\textsc{Distil (Anl\&Rw,1M)} & 19.45 & 4.03 & 13.52 && 11.40 & 1.55 & 19.17 && 15.46 & 6.87 & 12.45 && 21.59 & 8.42 & 21.96 \\
\bottomrule
\end{tabular}
\caption{
The proportion of three types of errors in the predictions.
}
\label{tab:error}
\end{table*}

Besides, we investigate the impact of the student model's size. Figure \ref{fig:scaling} illustrates that as the number of model parameters increases, the benefits of our distillation process exhibit a diminishing trend.
We attribute this trend to the decreasing gap in model size between the student and teacher, which narrows their gap in FSA capabilities and, consequently, reduces the potential gains from distillation.

\subsubsection{Results on Data-scarce Scenarios}
A major challenge of FSA is the relatively high cost of data annotation, which often leads to a scarcity of labeled data. 
We explore the feasibility of using distillation to mitigate this issue. 
Figure \ref{fig:scarce} indicates that the distilled model requires less than 25\% of the finetuning data to achieve the same results that the original SLM attains with full data.
Furthermore, we observe that the less finetuning data, the more significant the advantage of the distilled model over the original SLM, with improvements increasing from 6.00\% to 11.37\%.  
These findings suggest that distilling sentiment understanding from LLMs is a promising solution to the data scarcity challenge.

\subsection{Error Analysis}
\label{sec:error-analysis}

\citet{wang2024chatgpt} note that LLMs tend to produce plausible but not entirely accurate results. Inspired by this observation, we conduct an analysis of the wrong predictions. We first formulate three types of errors.
\textbf{Type 1 errors} occur when labels and predictions differ, but both are generally reasonable. For example, consider the review ``\textit{very hard to clean the dust off as there are multiple in-between segments, from metal to plastic to speakers}.'' 
In this case, categorizing the user opinion as either concerning \textsc{laptop\#usability} or \textsc{laptop\#design\_features} is acceptable.
\textbf{Type 2 errors} arise when insufficient context prevents the model from making accurate predictions. A typical example is ``\textit{I went through the settings and there isn't a way to fix it}.''
Without specific information about what `it' refers to, accurately inferring the appropriate aspect category becomes impossible.
\textbf{Type 3 errors} refer to errors other than the above two types, more fundamentally reflecting model inadequacy.

We sample about 100 wrong predictions and manually categorize them into these three types of errors, as shown in Table \ref{tab:error}.
We make the following observations.
(1) We observe  a notably high proportion of Type 1 errors, suggesting that the predictions are far more reasonable than exact-match metrics might indicate. This finding implies that as a subjective task, FSA requires evaluation metrics beyond exact matching, and relying solely on single labels could potentially hinder the model’s optimization.
(2) The proportion of Type 2 errors is also considerable, largely due to FSA annotations being at the sentence level. Conducting FSA at the review level could alleviate this issue.
(3) In Type 3 errors, LLMs significantly outperform SLMs, and distillation enhances the performance of SLMs.
In contrast, the first two types of errors cannot be reduced simply by enhancing the models. We look forward to future efforts to solve these issues through innovations in task formulations and metrics.

\section{Related Work}

\subsection{Pretraining for Sentiment Analysis}

Early studies \cite{xu-etal-2019-bert, gururangan-etal-2020-dont} show that pretraining on sentiment-specific corpora can significantly improve model performance in downstream sentiment analysis tasks. Subsequent works \cite{tian-etal-2020-skep, ke-etal-2020-sentilare, zhou-etal-2020-sentix, li-etal-2021-learning-implicit, fan-etal-2022-sentiment, SGPT} further integrate sentiment knowledge into the pretraining phase to encourage the model to learn sentiment-aware representations. This sentiment knowledge includes aspect words, sentiment words, review ratings, and emoticons. Additionally, some studies \cite{ke-etal-2020-sentilare,yin-etal-2020-sentibert, zhang-etal-2023-empirical} also incorporate syntactic knowledge during pretraining.

\subsection{Knowledge Distillation from LLMs}
Knowledge distillation \cite{hinton2015distilling} is the technique used to transfer knowledge from the teacher model to the student model.
It is most commonly applied in natural language processing for model compression \cite{sun-etal-2019-patient,sanh2020distilbert,jiao-etal-2020-tinybert,sun-etal-2020-mobilebert,NEURIPS2020_6f5216f8}.
With the advent of LLMs, many works explore transferring knowledge from large language models to existing SLMs to enhance their capabilities.
This research can be categorized into two main approaches.
The first leverages LLMs to generate pseudo-labels \cite{doi:10.1073/pnas.2305016120,ding-etal-2023-gpt,he2024annollm} or synthesize data \cite{NEURIPS2022_0346c148, ye-etal-2022-zerogen,he-etal-2023-targeted} and then use this data to train SLMs to equip them with the capabilities for specific tasks.
The second approach employs knowledge distillation to enable SLMs to enhance their reasoning capabilities \cite{ho-etal-2023-large, NEURIPS2023_97faedc9, wang-etal-2023-scott} or to generate rationales \cite{li-etal-2023-distilling}.

\section{Conclusions}

This paper explores the distillation from large language models (LLMs) for fine-grained sentiment analysis (FSA). We leverage the analysis and rewriting prompts to guide LLMs in generating sentiment understanding texts and then utilize these texts to pretrain small language models (SLMs). Additionally, we develop a comprehensive FSA benchmark to evaluate both SLMs and LLMs. Experimental results on this benchmark indicate that distillation significantly enhances the sentiment analysis capabilities of SLMs, yielding notable enhancements in both fully-supervised and zero-shot settings. This finding suggests that distillation is a highly promising direction for FSA. Moreover, our error analysis reveals that a considerable proportion of errors are not attributable to the model's capability, highlighting the need for innovative task formulations and metrics.

\section*{Limitations}

We list the potential limitations of this paper:
\begin{itemize}
\item 
While our distillation can reduce the inference cost of large language models (LLMs), it requires a large substantial of data and incurs high training costs. 
Exploiting logits, attention scores, and hidden states from LLMs may enhance distillation effectiveness and reduce data requirements.
\item Our error analysis in §\ref{sec:error-analysis} highlights the inadequacies of current exact-match metrics for subjective tasks such as fine-grained sentiment analysis. However, this paper does not offer a specific solution.
\end{itemize}
We believe that these limitations offer promising directions for future research.

\newpage
\bibliography{my,aaai25,anthology}

\cleardoublepage

\appendix

\begin{table}[ht]
\centering
\fontsize{8.8pt}{0.8\baselineskip}\selectfont
\setlength\tabcolsep{1pt}

\begin{tabular}{p{7.5cm}}
\toprule
\textbf{Analysis Prompt}

Analyze the fine-grained sentiment of a user review, listing both the opinion targets and the corresponding sentiments. 

\vspace{4pt}
Requirements:

- Opinion Target: Identify and locate the explicit mention of the reviewed entity within the review. 

- Aspect: Specify possible aspects mentioned on the opinion target, including entity types and attribute labels.

- Sentiment: Assess the sentiment intensity on the opinion target, selecting from very negative, negative, mild sentiment, positive, and very positive.

- Reasoning: Highlight the specific expressions of sentiment found within the review that are directed toward the opinion target. When sentiment is conveyed implicitly, explain how the sentiment is inferred from the context. An implicit sentiment does not utilize clear sentiment words or phrases (e.g., "happy," "disappointed," "love," or "hate") but rather is implied through context, including factual statements, comparisons, metaphors, or other indirect language.

\vspace{4pt}
Example:

\vspace{4pt}
\textcolor{blue}{\{demo\}}

\vspace{4pt}
Your Task:

\vspace{4pt}
Review: \textcolor{blue}{\{input review\}} \\
\midrule
\textbf{Rewriting Prompt}

Rewrite a user review to express the feelings more clearly and evidently. Incorporate descriptions of the evaluated aspects and highlight the corresponding sentiments. When sentiments are expressed implicitly, clarify them with direct assessments and explicitly state the feelings.

\vspace{4pt}

Example:

\vspace{4pt}
\textcolor{blue}{\{demo\}}

\vspace{4pt}
Your Task:

\vspace{4pt}
Review: \textcolor{blue}{\{input review\}} \\
\bottomrule
\end{tabular}
\caption{
The analysis and rewriting prompts.
}
\label{tab:analysis-rewriting-prompt}
\end{table}

\begin{table}[ht]
\centering
\fontsize{8.8pt}{0.8\baselineskip}\selectfont
\setlength\tabcolsep{1pt}
\begin{tabular}{p{7.5cm}} 
\toprule
\textbf{Targeted Sentiment Analysis}\\
Please perform targeted sentiment analysis task. Given the sentence, tag all (target, sentiment) pairs. Target should be substring of the sentence, and sentiment should be selected from [`negative', `neutral', `positive', `conflict']. If there are no target-sentiment pairs, return an empty list. Otherwise return a python list of tuples containing two strings in double quotes. Please return python list only, without any other comments or texts.

\vspace{4pt}
Sentence: I've been several times and am totally smitten.\\
Label: []\\
Sentence: The wine list is also really nice.\\
Label: [('wine list', 'positive')]\\

\vspace{4pt}
Sentence: I have to say they have one of the fastest delivery times in the city.\\
Label:\\
\midrule
\textbf{Aspect-level Sentiment Analysis}\\
Please perform aspect-level sentiment analysis task. Given the sentence, tag all (aspect category, sentiment) pairs. Aspect category should be selected from 
\textcolor{blue}{\{category space\}}, and sentiment should be selected from [`negative', `neutral', `positive']. If there are no target-sentiment pairs, return an empty list. Otherwise return a python list of tuples containing two strings in double quotes. Please return python list only, without any other comments or texts.

\vspace{4pt}
Sentence: so delicious!!!!!!\\
Label: [('food quality', 'positive')]\\
Sentence: The food arrived 20 minutes after I called, cold and soggy.\\
Label: [('food quality', 'negative'), ('service general', 'negative')]\\

\vspace{4pt}
Sentence: Serves really good sushi.\\
Label:\\
\bottomrule
\end{tabular}
\caption{
The prompts for in-context learning FSA.
}
\label{tab:icl-prompt}
\end{table}

\section*{Organization of Appendices}

We structure the appendix into four sections:
\begin{itemize}
\item Appendix A presents the complete prompts utilized in our paper;
\item Appendix B describes detailed annotations of fine-grained sentiment analysis (FSA) datasets and human evaluations;
\item Appendix C provides detailed hyperparameters for pretraining; and
\item Appendix D lists additional experimental results.
\end{itemize}

\section{Prompt Details}

\subsection{Analysis and Rewriting Prompts}
\label{app:analysis-rewriting-prompt}

We develop two prompts that guide large language models (LLMs) in generating content that embodies advanced sentiment understanding.
The first, the analysis prompt, aims to instruct LLMs to analyze the given review, listing the opinion targets and their corresponding aspects, sentiments, and reasonings. 
The second, the rewriting prompt, aims to instruct LLMs to rewrite the given review from a first-person perspective, expressing the user's feelings more clearly and evidently.
Table \ref{tab:analysis-rewriting-prompt} presents the complete analysis and rewriting prompts.

\begin{figure*}[t]
\centering
\includegraphics[width=0.88\linewidth]{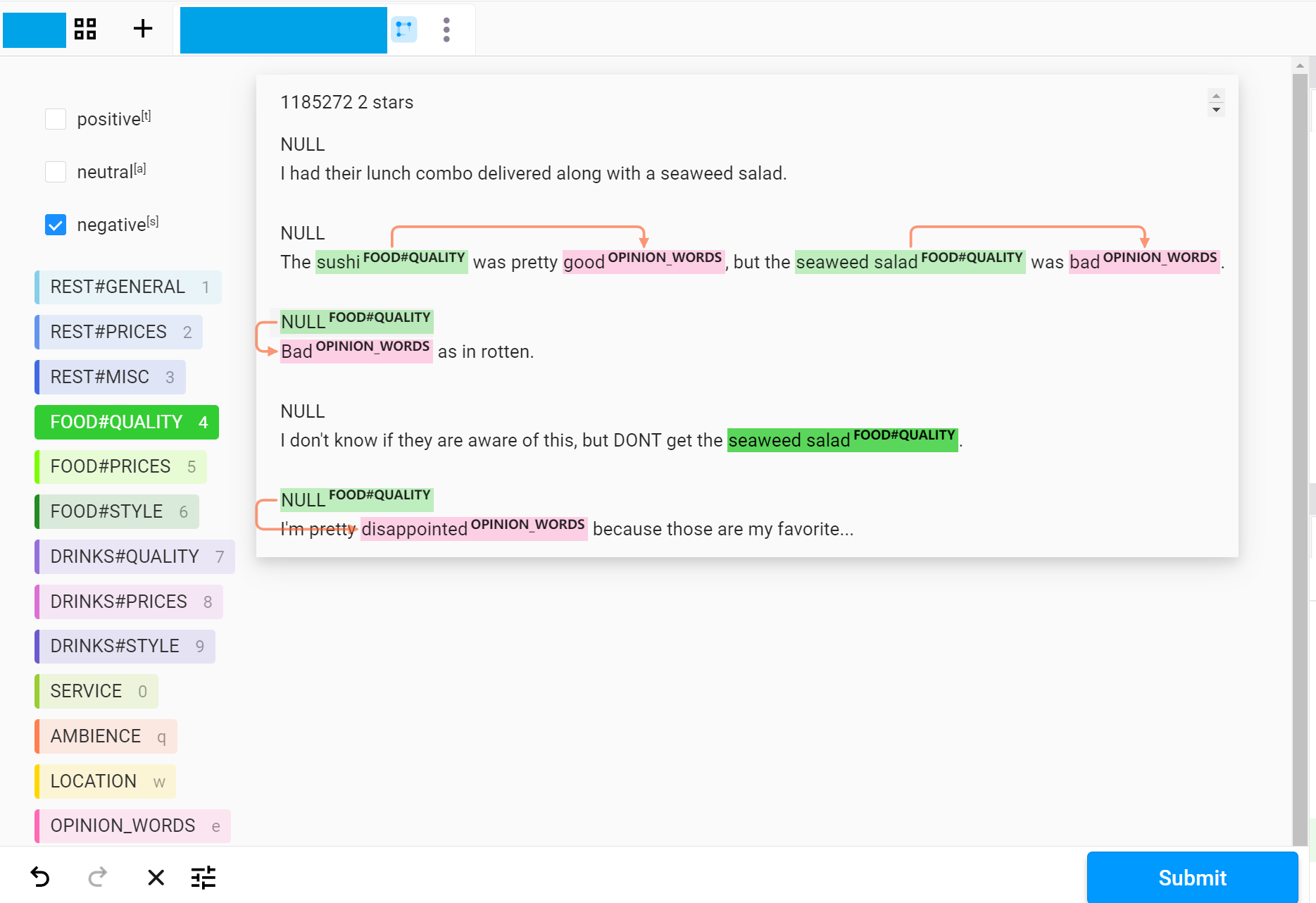}
\caption{
The screenshot of the annotation platform for FSA annotations.
}
\label{fig:6}
\end{figure*}

\subsection{In-context Learning Prompt}
\label{app:in-context-learning-prompt}

We examine the performance of LLMs on FSA datasets via in-context learning and fine-tuning.
Our in-context learning prompts follow \citet{zhang2023sentiment}, with the complete prompts provided in Table \ref{tab:icl-prompt}. 
The demonstration examples within the prompt are randomly sampled from the training set. We experimentally determine the number of demonstration examples: for TSA-Laptop14, it is set to 4; for other datasets, it is 8.

\subsection{Zero-shot Sentiment Classification Prompt}
\label{app:zero-shot-prompt}

\begin{table}[ht]
\centering
\fontsize{8.8pt}{0.8\baselineskip}\selectfont
\setlength\tabcolsep{1pt}
\begin{tabular}{p{7.5cm}} 
\toprule
\textbf{Targeted Sentiment Classification (GPT-3.5)}\\
Please perform the targeted sentiment classification task. Given the sentence, assign a sentiment label towards the opinion target from [`negative', `neutral', `positive', `conflict'].

Sentence: \textcolor{blue}{\{sentence\}}\\
Opinion target: \textcolor{blue}{\{target\}}\\
Label: \\
\midrule
\textbf{Aspect-level Sentiment Classification (GPT-3.5)}\\
Please perform the aspect-level sentiment classification task. Given the sentence, assign a sentiment label towards the aspect category from [`negative', `neutral', `positive'].

Sentence: \textcolor{blue}{\{sentence\}}\\
Aspect category: \textcolor{blue}{\{target\}}\\
Label: \\

\midrule

\textbf{Targeted Sentiment Classification (T5, Llama, Mixtral)}\\
Sentence: \textcolor{blue}{\{sentence\}}\\
Opinion target: \textcolor{blue}{\{target\}}\\
Label: \\
\midrule
\textbf{Aspect-level Sentiment Classification (T5, Llama, Mixtral)}\\
Sentence: \textcolor{blue}{\{sentence\}}\\
Aspect category: \textcolor{blue}{\{target\}}\\
Label: \\
\bottomrule
\end{tabular}
\caption{
The prompts for zero-shot sentiment classification.
}
\label{tab:zero-shot-prompt}
\end{table}

We also compare LLMs and small language models (SLMs) in zero-shot sentiment classification.
The prompts are presented in Table \ref{tab:zero-shot-prompt}.
For SLMs and open-source LLMs, Llama-2-7b and Mixtral-8x7b, we use simple prompts and determine the predictions by comparing the conditional probabilities of label words. For proprietary LLMs, GPT-3.5, we include an additional instruction.

\section{Annotation Details}
\subsection{FSA Annotation}
\label{app:fsa-annotation}

The test sets of original FSA datasets contain two few `hard samples', reducing the reliability of the evaluations.
Therefore, we annotate additional hard samples.
We provide the annotation details as follows.

\vspace{6pt}
\noindent
\textbf{Annotators.} The annotators comprise two well-educated students. One is a Ph.D. student specializing in sentiment analysis, with prior experience in annotating two FSA projects and extensive research in this field. The other is a master’s student researcher in fine-grained sentiment analysis. Both are well-versed in the concepts and annotation guidelines of FSA.

\begin{figure*}[t]
\centering
\includegraphics[width=1.\linewidth]{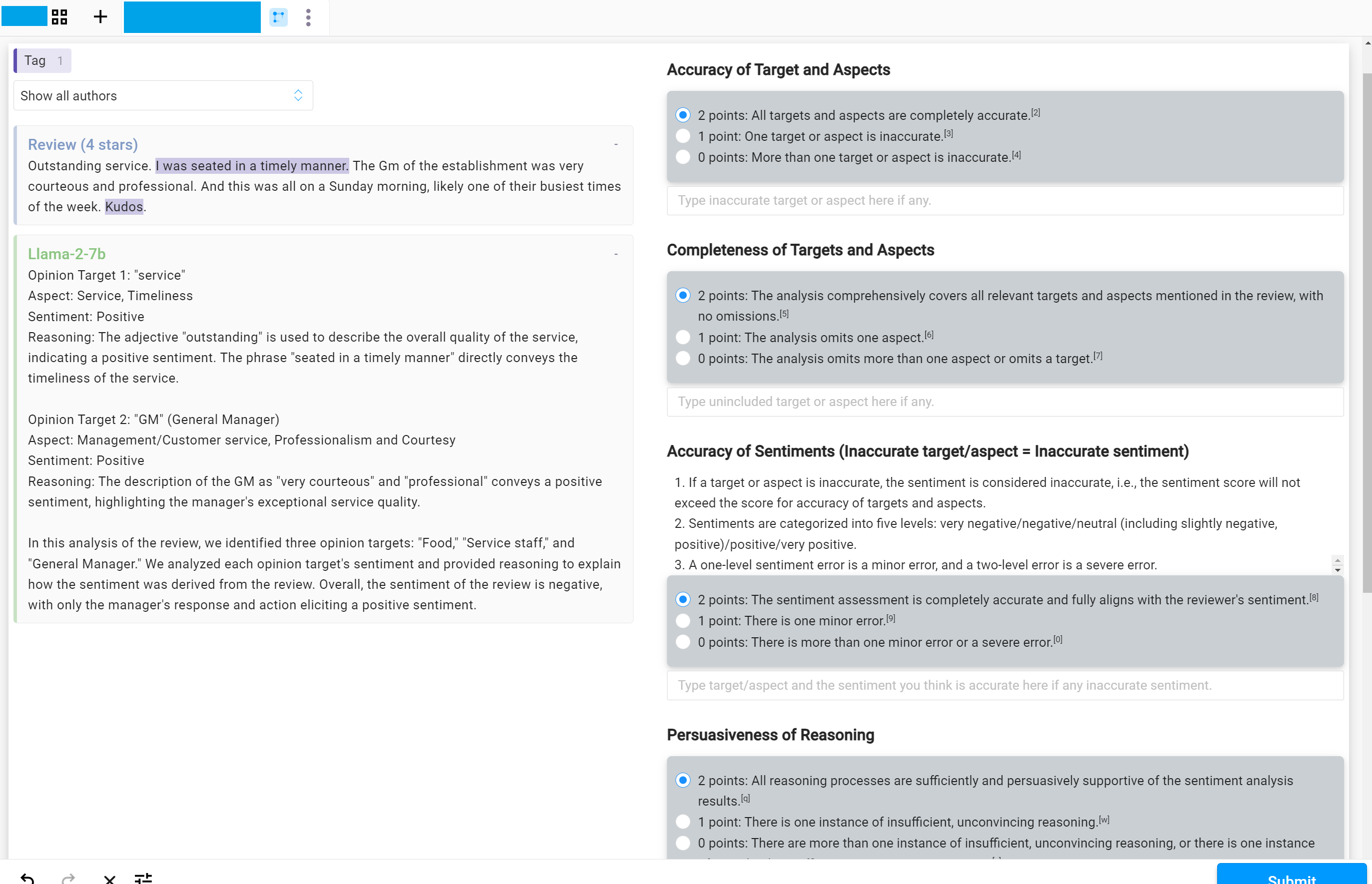}
\caption{The screenshot of the annotation platform for human evaluation.
}
\label{fig:7}
\end{figure*}

\vspace{6pt}
\noindent
\textbf{Annotation Guidelines.}
To ensure consistency, we provide annotators with well-developed and authoritative annotation guidelines \cite{pontiki-etal-2016-semeval}, which are available at \url{https://alt.qcri.org/semeval2016/task5/data/uploads/absa2016_annotationguidelines.pdf}.

\vspace{6pt}
\noindent
\textbf{Annotation Platform.}
We use Label Studio\footnote{\url{https://labelstud.io/}} to build our annotation platform, as shown in Figure \ref{fig:6}.
We begin by annotating the opinion targets and corresponding aspect categories within sentences. If aspect categories are inferred from the context without explicit text segments, the opinion targets are marked as `NULL'. Subsequently, we annotate the opinion words linked to each opinion target. Finally, we assess the corresponding sentiment.

\vspace{6pt}
\noindent
\textbf{Annotation Process.}
We collect 300 reviews each from the Yelp dataset\footnote{\url{https://www.yelp.com/dataset}} and the Amazon laptop dataset\footnote{\url{https://nijianmo.github.io/amazon/index.html}} \cite{ni-etal-2019-justifying}. The entire annotation process is divided into multiple batches, with each batch containing approximately 40 reviews. To ensure the quality of the annotations, each review is independently annotated by two annotators. At the end of each batch, the annotators will meet to discuss any inconsistencies based on the provided annotation guidelines. Data that cannot be reconciled will be removed.

\subsection{Human Evaluation for LLMs}
\label{app:human-evaluation}

We develop an evaluation framework to assess the sentiment understanding capabilities of LLMs.
We provide the annotation details as follows.

\vspace{6pt}
\noindent
\textbf{Annotators.}
The annotation team consists of four well-educated students, including one Ph.D. student, two master’s students, and one undergraduate student. 
All of them are well-versed in the provided annotation guidelines.

\vspace{6pt}
\noindent
\textbf{Annotation Guidelines.}
Our evaluation framework includes six metrics.
Each metric is scored from 0 to 2.
We develop detailed annotation guidelines for each metric to ensure consistency.
They are presented as follows.

\begin{center}
    \fcolorbox{black}{gray!10}{\parbox{0.94\linewidth}{
     Precision of Target and Aspects:\\
- 2 points: All targets and aspects are completely accurate.\\
- 1 point: One target or aspect is inaccurate.\\
- 0 points: More than one target or aspect is inaccurate.
    }}
\end{center}

\begin{center}
    \fcolorbox{black}{gray!10}{\parbox{.94\linewidth}{
     Recall of Targets and Aspects:\\
- 2 points: The analysis comprehensively covers all relevant targets and aspects mentioned in the review, with no omissions.\\
- 1 point: The analysis omits one aspect.\\
- 0 points: The analysis omits more than one aspect or omits a target.
    }}
\end{center}

\begin{center}
    \fcolorbox{black}{gray!10}{\parbox{.94\linewidth}{
     Accuracy of Sentiments:\\
     1) If a target or aspect is inaccurate, the sentiment is considered inaccurate, i.e., the sentiment score will not exceed the score for accuracy of targets and aspects.
2) Sentiments are categorized into five levels: very negative/negative/neutral (including slightly negative, positive)/positive/very positive.
3) A one-level sentiment error is a minor error, and a two-level error is a severe error.\\
- 2 points: The sentiment assessment is completely accurate and fully aligns with the reviewer's sentiment.\\
- 1 point: There is one minor error.\\
- 0 points: There is more than one minor error or a severe error.
    }}
\end{center}

\begin{center}
    \fcolorbox{black}{gray!10}{\parbox{.94\linewidth}{
     Persuasiveness of Reasoning:\\
- 2 points: All reasoning processes are sufficiently and persuasively supportive of the sentiment analysis results.\\
- 1 point: There is one instance of insufficient, unconvincing reasoning.\\
- 0 points: There are more than one instance of insufficient, unconvincing reasoning, or there is one instance of completely insufficient, unconvincing reasoning.
    }}
\end{center}

\begin{center}
    \fcolorbox{black}{gray!10}{\parbox{.94\linewidth}{
     Exhaustiveness of Reasoning:\\
- 2 points: All reasoning comprehensively lists all relevant sentiment segments, with no omissions.\\
- 1 point: One sentiment segment is omitted.\\
- 0 points: More than one sentiment segment is omitted.
    }}
\end{center}

\begin{center}
    \fcolorbox{black}{gray!10}{\parbox{.94\linewidth}{
     Hallucination Issues in Reasoning:\\
- 2 points: All reasoning does not use segments that do not exist in the original text, or uses irrelevant segments as evidence.\\
- 1 point: One instance.\\
- 0 points: More than one instance.
    }}
\end{center}

\vspace{6pt}
\noindent
\textbf{Annotation Platform.}
We use Label Studio to build our annotation platform, as shown in Figure \ref{fig:7}.
To ensure the quality of the annotations, we require annotators to identify the relevant segments before assigning scores.

\vspace{6pt}
\noindent
\textbf{Annotation Process.}
We collect 100 reviews each from the Yelp dataset and the Amazon laptop dataset. Then, we leverage the analysis prompt to guide three large language models in generating sentiment understanding texts. Each data is independently annotated by two annotators. We use the average of their scores as the final score.

\section{Implementation Details of Pretraining}
\label{app:implementation-details}

We continue to pretrain T5 on our sentiment understanding corpus.
Following \citet{nawrot2023nanot5}\footnote{\url{https://github.com/PiotrNawrot/nanoT5}}, we employ AdamW with RMS scaling as the optimization algorithm. 
The detailed hyperparameters are listed in Table \ref{tab:hp}.
Below, we describe how we determine the values of these hyperparameters.
\begin{itemize}
    \item The batch size is set to 100 based on our experience.
\item Learning rate and training epochs are determined experimentally. We try numerous combinations and find that the current combination (learning rate = 3e-3, training epochs = 10) achieves optimal performance.
\item Maximum seq lengths are decided by analyzing the length distribution of the pretraining data.
\item Other hyperparameters are adopted from \citet{nawrot2023nanot5}, which reproduces the pretraining of T5 in the PyTorch framework.
\end{itemize}

\begin{table}[t]
\centering
\fontsize{8.5pt}{0.8\baselineskip}\selectfont
\begin{tabular}{lc} 
\toprule
\textbf{Hyper-parameter} & \textbf{Value} \\ 
\midrule
Batch Size & 100 \\ 
Learning Rate & 3e-3 \\ 
Learning Rate Deacy & Cosine (fine\_cosine=1e-5) \\ 
Training Epoch & 10 \\ 
Warmup Step & 0 \\ 
Weight Decay & 0 \\ 
Adam $\beta_1$ & 0.9 \\ 
Adam $\beta_2$ & 0.999 \\ 
Gradient Clipping & 1.0 \\ 
Max Seq Length (Input) & 128 \\ 
Max Seq Length (Output) & 400 \\ 
\bottomrule
\end{tabular}
\caption{
Hyper-parameters for pretraining.
}
\label{tab:hp}
\end{table}

\section{Additional Results}

\begin{table}[t]
\centering
\fontsize{8.5pt}{.8\baselineskip}\selectfont
\setlength\tabcolsep{0.5pt}
\begin{tabular}{p{3.3cm} >{\centering\arraybackslash}p{1cm} >{\centering\arraybackslash}p{1.1cm} >{\centering\arraybackslash}p{1cm} >{\centering\arraybackslash}p{1.1cm} } 
\toprule
\textbf{Methods} & \textbf{Rest14} & \textbf{Laptop14} & \textbf{Rest16} & \textbf{Laptop16}\\
\midrule
T5-base            & 75.77 & 68.35 & 74.70 & 51.63 \\ 
DAPT               & 76.46 & 69.31 & 76.85 & 51.95 \\
Rating Prediction  & 77.52 & 68.46 & 75.78 & 52.77 \\
\midrule
\textbf{Llama-2-7b}          & \\
\textsc{In-context Learning} & 43.07 & 18.44 & 46.42 & 19.14 \\
\textsc{Supervised Finetuning}          & 78.74 & 71.23 & 79.59 & 60.13 \\
\rowcolor{gray!20} 
\textsc{Distil (ANL\&Rw)}    & 77.15 & 71.09 & 77.99 & 53.48 \\
\rowcolor{gray!20} 
\textsc{Distil (ANL\&Rw, 1M)}& 78.06 & 71.73 & 79.06 & 55.98\\
\midrule
\textbf{Mixtral-8x7b}\\
\textsc{In-context Learning} & 53.54 & 29.74 & 60.56 & 29.53\\
\textsc{Supervised Finetuning} & 80.40 & 72.44 & 82.91 & 62.54 \\
\rowcolor{gray!20} 
\textsc{Distil (ANL\&Rw)} & 77.83 & 72.61 & 77.64 & 54.21 \\
\rowcolor{gray!20} 
\textsc{Distil (ANL\&Rw, 1M)} & 79.05 & 73.48 & 79.41 & 56.89 \\
\midrule
\textbf{GPT-3.5}\\
\textsc{In-context Learning} & 55.36 & 27.08 & 61.56 & 27.99\\
\rowcolor{gray!20} 
\textsc{Distil (ANL\&Rw)} & 77.68 & 71.03 & 78.14 & 54.33 \\
\bottomrule
\end{tabular}
\caption{
Experimental results on the original test sets ($F_1$-score, \%).
}
\label{tab:11}
\end{table}

\subsection{Results on Original Test Sets}
\label{app:additional-results}

To facilitate comparisons for future work, we also publish the performance on the original FSA test sets in Table \ref{tab:11}.

\subsection{Results on the ASQP Task}

Aspect sentiment quad prediction (ASQP) \cite{cai-etal-2021-aspect,zhang-etal-2021-aspect-sentiment} is a comprehensive task in FSA that has received widespread attention. We also conduct experiments on this task, and the results are displayed in Table \ref{tab:asqp-result}.
We can see that our distillation also significantly enhances the performance of SLMs on this task.

\begin{table}[t]
\centering
\fontsize{8.5pt}{0.8\baselineskip}\selectfont
\setlength\tabcolsep{2.7pt}
\begin{tabular}{l ccccc} 
\toprule
\multirow{2}*{\textbf{Methods}}  & \multicolumn{2}{c}{\textbf{ACOS}} &\multicolumn{2}{c}{\textbf{ASQP}} & \multirow{2}*{\textbf{Avg}} \\
\cmidrule(lr){2-3} \cmidrule(lr){4-5}
& {Laptop16} & {Rest16} & {Rest15} 
& {Rest16}  \\
\midrule
T5-base & 44.35 & 59.77 & 48.23 & 59.17 & 52.88\\
\hdashline[2pt/4pt]
\textsc{Distil} (Llama) & 44.68 & 62.00 & 51.78 & 62.97 & +2.48\\ 
\textsc{Distil} (Mixtral) & 45.44 & 62.41 & 52.99 & 62.87 & +3.05\\ 
\textsc{Distil} (GPT-3.5) & 45.21 & 63.09 & 50.38 & 61.95 & +2.28\\ 
\bottomrule
\end{tabular}
\caption{
Experimental results on four ASQP datasets ($F_1$-score, \%).
Distillation for Llama-2-7b and Mixtral-8x7b leverage 1 million reviews, whereas GPT-3.5 leverages 100K reviews.
}
\label{tab:asqp-result}
\end{table}

\end{document}